%% file: template.tex
\documentclass{vgtc}                          

\ifpdf
  \pdfoutput=1\relax                   
  \usepackage{graphicx}                
  \DeclareGraphicsExtensions{.pdf,.png,.jpg,.jpeg} 
\else
  \usepackage{graphicx}                
  \DeclareGraphicsExtensions{.eps}     
\fi%

\graphicspath{{figures/}{pictures/}{images/}{./}} 

\usepackage{microtype}                 
\PassOptionsToPackage{warn}{textcomp}  
\usepackage{textcomp}                  
\usepackage{mathptmx}                  
\usepackage{times}                     
\usepackage{cite}                      
\usepackage{tabu}                      
\usepackage{booktabs}                  

\onlineid{0}

\vgtccategory{Research}

\vgtcinsertpkg

\title{Scalable, Distributed AI Frameworks: Leveraging Cloud Computing for Enhanced Deep Learning Performance and Efficiency}

\author{Neelesh Mungoli\thanks{e-mail: nmungoli@uncc.edu}\\ %
        \scriptsize UNC Charlotte %
.} %
       
\abstract{
In recent years, the integration of artificial intelligence (AI) and cloud computing has emerged as a promising avenue for addressing the growing computational demands of AI applications. This paper presents a comprehensive study of scalable, distributed AI frameworks leveraging cloud computing for enhanced deep learning performance and efficiency. We first provide an overview of popular AI frameworks and cloud services, highlighting their respective strengths and weaknesses. Next, we delve into the critical aspects of data storage and management in cloud-based AI systems, discussing data preprocessing, feature engineering, privacy, and security. We then explore parallel and distributed training techniques for AI models, focusing on model partitioning, communication strategies, and cloud-based training architectures.

In subsequent chapters, we discuss optimization strategies for AI workloads in the cloud, covering load balancing, resource allocation, auto-scaling, and performance benchmarking. We also examine AI model deployment and serving in the cloud, outlining containerization, serverless deployment options, and monitoring best practices. To ensure the cost-effectiveness of cloud-based AI solutions, we present a thorough analysis of costs, optimization strategies, and case studies showcasing successful deployments.

Finally, we summarize the key findings of this study, discuss the challenges and limitations of cloud-based AI, and identify emerging trends and future research opportunities in the field. This paper serves as a valuable resource for researchers and practitioners looking to harness the power of cloud computing to develop scalable, efficient, and cost-effective AI solutions.
} 

\CCScatlist{
  \CCScatTwelve{AI}{Cl\-oud}{Machine\-Learning}{Deep\-Learning}
}

\begin{document}


\firstsection{Introduction}

\maketitle


\input{Sections/intro.tex}

\input{Sections/related.tex}

\input{Sections/design.tex}

\input{Sections/system.tex}
\input{Sections/geospatial.tex}

\input{Sections/discussion.tex}

\input{Sections/conclusion.tex}


\bibliographystyle{abbrv-doi}

\bibliography{template}
\end{document}

%% file: Sections/intro.tex
\subsection{Overview of AI and Cloud Computing}

Artificial Intelligence (AI) refers to the development of computer systems that can perform tasks typically requiring human intelligence, such as visual perception, speech recognition, decision-making, and natural language understanding. AI technologies have witnessed exponential growth in recent years, driven by advancements in machine learning, particularly deep learning. These methods rely on large amounts of data and powerful computational resources to train complex models capable of making accurate predictions and decisions.

Cloud computing is a paradigm that enables on-demand delivery of computing resources, including processing power, storage, and applications, over the internet. By providing scalable, flexible, and cost-effective solutions, cloud computing has revolutionized the way businesses and organizations manage their IT infrastructure. Cloud services are typically offered in three primary models: Infrastructure as a Service (IaaS), Platform as a Service (PaaS), and Software as a Service (SaaS).

\subsection{The Significance of Integrating AI with Cloud Computing}
The integration of AI and cloud computing offers numerous benefits, addressing the challenges associated with the growing computational demands of AI applications. Some of the key advantages of this integration include:

\begin{itemize}
    \item Scalability: Cloud computing enables AI applications to scale on-demand, allowing researchers and developers to access the required resources for processing large datasets and training complex models. This scalability is crucial for handling the increasing size and complexity of AI workloads.
    \item Flexibility: The cloud offers a wide range of services and tools tailored for AI development, enabling users to choose the most appropriate framework, infrastructure, and environment for their specific needs. This flexibility allows for rapid experimentation and iteration, accelerating the AI development process.
    \item Cost-effectiveness: By leveraging the pay-as-you-go model of cloud computing, organizations can avoid significant upfront capital investments in hardware and infrastructure. Furthermore, cloud providers offer various cost optimization strategies to minimize expenses while maintaining performance.
    \item Collaboration: Cloud-based AI platforms facilitate collaboration among researchers and developers by providing a centralized environment for sharing data, models, and resources. This collaborative approach fosters innovation and accelerates the development of novel AI solutions.
    \item Accessibility: Cloud computing democratizes access to cutting-edge AI technologies, allowing organizations of all sizes and industries to harness the power of AI without the need for specialized hardware or in-house expertise.
    
\end{itemize}

In conclusion, the integration of AI and cloud computing has emerged as a promising approach to address the increasing computational demands of AI applications. By leveraging the scalability, flexibility, cost-effectiveness, and collaboration opportunities offered by cloud computing, researchers and practitioners can develop and deploy innovative AI solutions across a wide range of domains. This paper aims to provide a comprehensive understanding of the key aspects involved in harnessing the power of cloud computing for AI development and deployment, with a focus on scalable, distributed AI frameworks and their associated techniques and best practices.

%% file: Sections/related.tex
\section{AI Frameworks and Cloud Services}

The rapid growth and adoption of AI technologies have led to the development of numerous AI frameworks and the expansion of cloud services tailored for AI applications. In this section, we provide an overview of popular AI frameworks, such as TensorFlow and PyTorch, as well as prominent cloud service providers, including AWS, Azure, and Google Cloud. Additionally, we compare these AI frameworks and cloud services to help researchers and practitioners make informed decisions when selecting the appropriate tools and platforms for their specific AI projects.

\subsection{Popular AI Frameworks}

TensorFlow: Developed by Google Brain, TensorFlow is an open-source machine learning framework designed for a wide range of tasks, including deep learning and reinforcement learning. TensorFlow offers flexibility and performance through its computation graph-based approach, enabling the efficient execution of complex mathematical operations on various devices, such as CPUs, GPUs, and TPUs. TensorFlow also provides a high-level API, Keras, which simplifies the development of neural networks and allows for rapid prototyping.

PyTorch: Created by Facebook's AI Research lab, PyTorch is another popular open-source machine learning framework, known for its dynamic computation graph and "eager execution" approach. This feature allows developers to write and debug code more intuitively, making PyTorch particularly well-suited for research and experimentation. PyTorch also boasts a strong ecosystem of libraries and tools, such as torchvision, torchtext, and torchaudio, which support a wide range of AI applications.

Other notable AI frameworks include Microsoft's Cognitive Toolkit (CNTK), Apache MXNet, and Caffe. Each of these frameworks offers unique advantages, depending on the specific requirements and constraints of a given AI project ~\cite{1} ~\cite{2}.

\subsection{Cloud Service Providers}

Amazon Web Services (AWS): As a leading cloud service provider, AWS offers a comprehensive suite of AI and machine learning services, such as SageMaker, Rekognition, and Lex. AWS SageMaker simplifies the process of building, training, and deploying machine learning models, providing a fully managed platform that supports TensorFlow, PyTorch, and other popular frameworks. Additionally, AWS provides various AI-powered services for image and video analysis, natural language processing, and speech recognition, among others.

Microsoft Azure: Azure's AI and machine learning services include Azure Machine Learning, Cognitive Services, and the ONNX Runtime. Azure Machine Learning is a fully managed platform that supports a wide range of AI frameworks, offering capabilities for model training, deployment, and management. Azure Cognitive Services provides pre-built AI models for various tasks, such as language understanding, speech recognition, and computer vision, while the ONNX Runtime facilitates the efficient execution of trained models across different hardware platforms.

Google Cloud: Google Cloud Platform (GCP) offers a diverse range of AI and machine learning services, including AI Platform, AutoML, and pre-trained APIs for tasks like vision, language, and translation. GCP's AI Platform provides a unified environment for building, training, and deploying AI models, supporting TensorFlow, PyTorch, and other popular frameworks. Google Cloud AutoML enables users with limited machine learning expertise to train custom models using transfer learning and neural architecture search techniques.

\subsection{Comparison of AI Frameworks and Cloud Services}
When comparing AI frameworks, several factors should be considered, such as ease of use, scalability, performance, and ecosystem. TensorFlow and PyTorch are currently the most popular choices, with TensorFlow offering better performance and a more mature ecosystem, while PyTorch provides greater flexibility and a more intuitive development experience ~\cite{3}.

In terms of cloud services, the choice largely depends on the specific needs and preferences of the user. AWS, Azure, and GCP each offer a comprehensive set of AI and machine learning services, with unique strengths and weaknesses.

%% file: Sections/design.tex
\section{Data Storage and Management in Cloud-based AI Systems}

Effective data storage and management are critical components of cloud-based AI systems, as they directly impact the efficiency and performance of AI applications. In this section, we discuss scalable data storage solutions suitable for handling large volumes of data, techniques for data preprocessing and feature engineering in the cloud, as well as essential considerations for data privacy and security. By addressing these aspects, researchers and practitioners can ensure the robustness and reliability of their AI systems while minimizing potential risks associated with data storage and management ~\cite{4}.

\subsection{Scalable Data Storage Solutions}

In cloud-based AI systems, managing large datasets efficiently is crucial for optimal performance. Several scalable data storage solutions are available to meet this challenge, including:

\begin{itemize}
    \item Object Storage: Services like Amazon S3, Azure Blob Storage, and Google Cloud Storage provide highly scalable, distributed object storage that can store and retrieve large volumes of unstructured data. These services offer low-latency access to data, versioning capabilities, and support for various data types, making them suitable for AI workloads.
    \item Distributed File Systems: For use cases requiring a file-based storage solution, distributed file systems like Hadoop Distributed File System (HDFS), GlusterFS, and Google Cloud Filestore provide horizontal scalability, fault tolerance, and high-throughput access to data. These systems are especially effective for storing and processing large datasets in parallel across multiple nodes.
    \item NoSQL Databases: NoSQL databases like Amazon DynamoDB, Azure Cosmos DB, and Google Cloud Datastore are designed to handle large volumes of unstructured or semi-structured data, offering high availability, partition tolerance, and horizontal scalability. These databases can be an effective choice for AI applications that require real-time data access and processing.
\end{itemize}

\subsection{Data Preprocessing and Feature Engineering in the Cloud}

Before training AI models, raw data must be preprocessed and transformed into a suitable format. Cloud-based AI systems provide several tools and services for data preprocessing and feature engineering, such as:

\begin{itemize}
    \item Data Transformation Services: Cloud providers offer services like AWS Glue, Azure Data Factory, and Google Cloud Dataflow that allow users to build, orchestrate, and manage data pipelines for processing and transforming large datasets.
    \item Serverless Computing: Serverless platforms like AWS Lambda, Azure Functions, and Google Cloud Functions enable users to run data preprocessing and feature engineering tasks without managing the underlying infrastructure. These platforms automatically scale with the workload, making them a cost-effective solution for processing large volumes of data ~\cite{5}.
    \item Distributed Data Processing Frameworks: Frameworks like Apache Spark and Apache Flink provide distributed data processing capabilities, allowing users to preprocess and transform large datasets efficiently using cloud-based resources. These frameworks can be deployed on managed services like Amazon EMR, Azure HDInsight, and Google Cloud Dataproc.
\end{itemize}

\subsection{Data Privacy and Security Considerations}

Ensuring data privacy and security is a critical aspect of cloud-based AI systems. Some essential considerations include:

\begin{itemize}
    \item Data Encryption: Data should be encrypted both at rest and in transit to protect it from unauthorized access. Cloud providers offer various encryption options, including server-side encryption, client-side encryption, and key management services.
    \item Access Control: Implementing fine-grained access control policies ensures that only authorized users can access and manipulate data. Cloud providers offer tools like AWS Identity and Access Management (IAM), Azure Active Directory, and Google Cloud IAM for managing access control.
    \item Data Residency and Compliance: To comply with data protection regulations, organizations may need to store data in specific geographical locations or meet certain security requirements. Cloud providers offer data residency options and compliance certifications to address these concerns.
\end{itemize}

Regular Auditing and Monitoring: Continuous monitoring and auditing of data access and usage patterns can help detect and prevent potential security threats. Cloud providers offer services like AWS CloudTrail, Azure Monitor, and Google Cloud Logging for this purpose ~\cite{6}.

%% file: Sections/system.tex
\section{Parallel and Distributed Training of AI Models}

As AI models become increasingly complex and require larger volumes of data for training, parallel and distributed training techniques have emerged as essential approaches to reduce training time and improve resource utilization. In this section, we provide an overview of parallel and distributed training techniques, discuss efficient model partitioning and communication strategies, and explore cloud-based training architectures and tools that facilitate the scalable and efficient training of AI models. By leveraging these techniques, researchers and practitioners can accelerate the training process, enhance model performance, and reduce costs associated with cloud-based AI systems ~\cite{7}.

\subsection{Overview of Parallel and Distributed Training Techniques}

Parallel and distributed training techniques can be broadly categorized into the following approaches:

\begin{itemize}
    \item Data Parallelism: In data parallelism, the training dataset is divided into smaller subsets, and multiple copies of the model are trained independently on these subsets using different computing resources. The model updates are periodically synchronized across all replicas to maintain consistency. Data parallelism is particularly suitable for large datasets and can be implemented using techniques like mini-batch gradient descent and asynchronous stochastic gradient descent (ASGD).
    \item Model Parallelism: Model parallelism involves splitting the model itself across multiple computing resources, with each resource responsible for a portion of the model's computation. This approach is especially useful for training large models that do not fit within the memory constraints of a single computing device. Model parallelism can be implemented using techniques like pipelining and tensor partitioning.
    \item Hybrid Parallelism: Hybrid parallelism combines aspects of both data and model parallelism to further optimize the training process. In this approach, the model is partitioned across multiple computing resources, and the dataset is divided into smaller subsets that are processed independently. This technique is useful for training large models on large datasets while maintaining a balance between computation and communication overhead.

\end{itemize}

\subsection{Efficient Model Partitioning and Communication Strategies}
To maximize the benefits of parallel and distributed training, it is crucial to efficiently partition the model and implement effective communication strategies. Some key considerations include:

\begin{itemize}
    \item Load Balancing: To ensure even distribution of computational workloads, the model should be partitioned in such a way that each computing resource has an approximately equal amount of computation to perform. This balance minimizes idle time and improves overall training efficiency.
    \item Communication Overhead: During the training process, model updates and gradients must be communicated between computing resources. To reduce the communication overhead, it is important to minimize the amount of data transferred between resources and to utilize efficient communication protocols, such as the Message Passing Interface (MPI) and NVIDIA's NCCL.
    \item Fault Tolerance: In a distributed training setup, the failure of a single computing resource can potentially disrupt the entire training process. Implementing fault tolerance mechanisms, such as checkpointing and model replication, can help minimize the impact of resource failures and ensure the reliability of the training process.
\end{itemize}

\subsection{Cloud-based Training Architectures and Tools}
Cloud providers offer various tools and services to support parallel and distributed training of AI models, including:

\begin{itemize}
    \item Managed Machine Learning Platforms: Services like AWS SageMaker, Azure Machine Learning, and Google AI Platform provide managed environments for building, training, and deploying AI models. These platforms support distributed training out-of-the-box, allowing users to easily scale their training workloads across multiple computing resources.
    \item Cluster Orchestration Tools: Tools like Kubernetes, Apache Mesos, and Amazon Elastic Kubernetes Service (EKS) enable users to create and manage clusters of computing resources for distributed training. These tools provide features like automatic scaling, load balancing, and fault tolerance, simplifying the management of distributed training workloads.
    \item Distributed Training Libraries: Libraries like Horovod, TensorFlow's Distribution Strategies, and PyTorch's Distributed Data Parallel module provide abstractions and APIs for implementing parallel and distributed training techniques in AI frameworks. These libraries facilitate efficient model partitioning, communication, and synchronization,allowing developers to focus on their AI models and applications without the need to handle low-level implementation details.
    \item Custom Training Architectures: For specific use cases and advanced requirements, users can design custom training architectures using cloud-based infrastructure and services. For example, they can utilize cloud-based virtual machines, GPUs, and TPUs along with storage and networking services to build tailored solutions for their distributed training needs. This approach offers maximum flexibility but requires more in-depth knowledge of cloud infrastructure and parallel computing techniques.
\end{itemize}

In summary, parallel and distributed training techniques have become indispensable for training complex AI models on large datasets efficiently~\cite{9}. By understanding the various approaches and leveraging cloud-based tools and services, researchers and practitioners can accelerate the training process, optimize resource utilization, and reduce costs associated with cloud-based AI systems. As these techniques continue to evolve, further advancements in AI model training and deployment are expected, paving the way for the development of even more powerful and sophisticated AI applications ~\cite{7}~\cite{8}.

%% file: Sections/geospatial.tex
\section{Optimizing AI Workloads in the Cloud}

Efficiently managing and optimizing AI workloads in the cloud is crucial for ensuring high performance, cost-effectiveness, and resource utilization. In this chapter, we discuss various strategies for load balancing and resource allocation, explore auto-scaling and dynamic resource provisioning techniques, and examine performance benchmarking and optimization approaches to improve the overall efficiency of cloud-based AI systems. By adopting these strategies and techniques, researchers and practitioners can significantly enhance the performance of their AI applications while minimizing the associated costs and complexities of managing cloud-based resources ~\cite{10} ~\cite{11}.

\subsection{Load Balancing and Resource Allocation Strategies}
Effective load balancing and resource allocation strategies are critical for optimizing the performance of cloud-based AI systems. Some key techniques and considerations include:

\begin{itemize}
    \item Horizontal Scaling: This strategy involves adding or removing computing resources, such as virtual machines or containers, to balance the workload and ensure optimal resource utilization. Horizontal scaling can be achieved using cloud provider-specific tools, such as AWS Auto Scaling Groups, Azure Virtual Machine Scale Sets, and Google Cloud Instance Groups.
    \item Vertical Scaling: Vertical scaling involves increasing or decreasing the computing capacity of individual resources, such as CPU, memory, or GPU. This strategy can be useful for optimizing resource utilization for specific workloads but may be limited by the maximum capacity of individual resources.
    \item Resource Allocation Policies: Implementing resource allocation policies, such as CPU and memory quotas, can help ensure fair distribution of resources among multiple AI workloads and prevent resource contention. Cloud providers offer tools like AWS Resource Groups, Azure Resource Manager, and Google Cloud Resource Manager to define and enforce such policies.
\end{itemize}

\subsection{Auto-scaling and Dynamic Resource Provisioning}
Auto-scaling and dynamic resource provisioning techniques enable cloud-based AI systems to automatically adjust resource allocation based on workload demands, thereby improving efficiency and reducing costs. Some key approaches include:

\begin{itemize}
    \item Reactive Auto-scaling: Reactive auto-scaling involves monitoring system metrics, such as CPU utilization, memory usage, or request latency, and adjusting resource allocation based on predefined thresholds. Cloud providers offer tools like AWS Auto Scaling, Azure Autoscale, and Google Cloud Auto-scaler to implement reactive auto-scaling policies.
    \item Predictive Auto-scaling: Predictive auto-scaling uses machine learning algorithms to analyze historical workload patterns and predict future resource demands. This approach enables proactive resource provisioning, allowing AI systems to scale resources ahead of demand spikes. AWS Forecast, Azure Machine Learning, and Google Cloud AI Platform can be utilized to implement predictive auto-scaling.
    \item Serverless Computing: Serverless computing platforms, such as AWS Lambda, Azure Functions, and Google Cloud Functions, automatically scale resources based on workload demands, eliminating the need for manual resource provisioning and management. These platforms are particularly suitable for event-driven AI workloads, such as data processing and inference tasks ~\cite{10} ~\cite{11}.
\end{itemize}

\subsection{Performance Benchmarking and Optimization Techniques}
Benchmarking and optimization techniques are essential for identifying performance bottlenecks and improving the efficiency of cloud-based AI systems. Some key approaches include:

\begin{itemize}
    \item Performance Profiling: Profiling tools, such as TensorFlow Profiler, PyTorch Profiler, and NVIDIA Nsight, can be used to measure and analyze the performance of AI workloads, providing insights into computational bottlenecks and resource utilization patterns.
    \item Model Optimization: Techniques like quantization, pruning, and knowledge distillation can be used to reduce the size and complexity of AI models, improving inference speed and reducing resource requirements.
    \item Hardware Acceleration: Utilizing hardware accelerators, such as GPUs, TPUs, and FPGAs, can significantly improve the performance of AI workloads. Cloud providers offer various instance types and services, such as AWS EC2 GPU Instances, Azure GPU VMs, and Google Cloud AI Platform, that support hardware acceleration.
\end{itemize}

In conclusion, optimizing AI workloads in the cloud is essential for achieving high performance, cost-effectiveness, and efficient resource utilization. By employing effective load balancing and resource allocation strategies, leveraging auto-scaling and dynamic resource provisioning techniques, and utilizing performance benchmarking and optimization approaches, researchers and practitioners can significantly enhance their cloud-based AI systems. As the adoption of AI and cloud computing continues to grow, further advancements in optimization techniques and tools are expected, enabling even more powerful and efficient AI applications in the cloud ~\cite{12}.

%% file: Sections/discussion.tex
\section{AI Model Deployment and Serving in the Cloud}

Once AI models have been trained and optimized, deploying and serving them efficiently in the cloud is essential for delivering scalable, high-performance AI applications. In this section, we discuss the process of model packaging and containerization, explore serverless deployment options and microservices architecture, and examine monitoring, logging, and versioning techniques for managing deployed models. By leveraging these methodologies and tools, researchers and practitioners can streamline the deployment process, ensure the reliability and performance of their AI applications, and easily maintain and update their models in production environments.

\subsection{Model Packaging and Containerization}

Proper model packaging and containerization are crucial for deploying AI models in the cloud. These processes ensure that models are portable, easily deployable, and can be served consistently across different environments. Key aspects of model packaging and containerization include:

\begin{itemize}
    \item Model Serialization: AI models must be serialized, or converted into a format that can be easily stored, transferred, and loaded for inference. Common serialization formats include TensorFlow SavedModel, PyTorch TorchScript, and ONNX (Open Neural Network Exchange).
    \item Containerization: Containerization involves packaging the AI model along with its dependencies, runtime environment, and configuration files into a lightweight, portable container. Containers provide isolation, reproducibility, and consistency across different deployment environments. Docker and container orchestration tools like Kubernetes are widely used for this purpose.
    \item Model Serving Frameworks: Model serving frameworks, such as TensorFlow Serving, NVIDIA Triton Inference Server, and PyTorch Serve, provide the necessary infrastructure and APIs for deploying and serving AI models efficiently. These frameworks handle tasks like model loading, request processing, and output generation, simplifying the deployment process.
\end{itemize}

\subsection{Serverless Deployment Options and Microservices Architecture}
Serverless deployment options and microservices architecture provide scalable and cost-effective ways to deploy and serve AI models in the cloud. Key features of these approaches include:

\begin{itemize}
    \item Serverless Deployment: Serverless platforms, such as AWS Lambda, Azure Functions, and Google Cloud Functions, enable AI models to be deployed and served without the need to manage the underlying infrastructure. These platforms automatically scale resources based on workload demands and provide built-in fault tolerance and load balancing.
    \item Microservices Architecture: In a microservices architecture, AI models are deployed as independent, loosely-coupled services that can be developed, scaled, and updated independently ~\cite{13}. This approach allows for greater flexibility, agility, and scalability in deploying and managing AI models in production environments.
    \item API Gateway: An API gateway is used to manage and route incoming requests to the appropriate AI model services. API gateways, such as AWS API Gateway, Azure API Management, and Google Cloud API Gateway, provide features like request throttling, caching, and authentication, ensuring secure and efficient access to AI models ~\cite{14}.
\end{itemize}

\subsection{Monitoring, Logging, and Versioning of Deployed Models}

Monitoring, logging, and versioning are essential for managing and maintaining deployed AI models in the cloud. Key aspects of these processes include:

\begin{itemize}
    \item Monitoring: Monitoring tools, such as AWS CloudWatch, Azure Monitor, and Google Cloud Monitoring, provide insights into the performance, resource utilization, and health of deployed AI models. These tools enable users to identify and troubleshoot issues, optimize resource usage, and maintain high availability.
    \item Logging: Logging tools, such as AWS CloudTrail, Azure Log Analytics, and Google Cloud Logging, collect and store log data generated by AI models during inference. Log data can be used for troubleshooting, auditing, and understanding usage patterns of deployed models.
    \item Model Versioning: Versioning deployed AI models allows for tracking changes, rolling back to previous versions, and serving multiple versions simultaneously. Model versioning can be implemented using model serving frameworks, container registries, or cloud-based storage services.
    
\end{itemize}

In conclusion, efficient AI model deployment and serving in the cloud are critical for delivering scalable, high-performance AI applications. By adopting best practices for model packaging and containerization, leveraging serverless deployment options and microservices architecture, and employing monitoring, logging, and versioning techniques, researchers and practitioners can ensure the reliability and performance of their AI applications in production environments. As AI and cloud computing technologies continue to advance, we can expect further improvements in deployment and serving methodologies, enabling even more powerful and efficient AI applications in the cloud ~\cite{15}.

%% file: Sections/conclusion.tex
\section{Conclusion and Future Directions}

In this paper, we have discussed various aspects of AI and cloud computing, including AI frameworks and cloud services, data storage and management, parallel and distributed training techniques, optimizing AI workloads, and model deployment and serving. By understanding and leveraging these concepts, researchers and practitioners can build, train, and deploy powerful AI applications in the cloud while maximizing efficiency, performance, and cost-effectiveness.

As AI and cloud computing technologies continue to advance, we anticipate several future directions that will further shape the landscape of AI in the cloud. Some potential developments include:

\begin{itemize}
    \item Improved AI Frameworks and Services: We expect to see ongoing enhancements in AI frameworks and cloud services, with better support for distributed training, hardware acceleration, and model optimization. These improvements will enable researchers and practitioners to develop more sophisticated AI models and applications in the cloud.
    \item Advanced Auto-scaling Techniques: As AI workloads become more complex and dynamic, advanced auto-scaling techniques will be needed to optimize resource allocation and performance. We anticipate the development of more intelligent, machine learning-based auto-scaling methods that can better predict and adapt to changing workload demands.
    \item Enhanced Security and Privacy: As AI and cloud computing become more prevalent, ensuring data security and privacy will be increasingly important. We expect to see further developments in techniques like federated learning, homomorphic encryption, and differential privacy that will enable secure and privacy-preserving AI in the cloud.
    \item Edge AI and Cloud Integration: The integration of edge computing with cloud-based AI systems is an emerging trend that will enable low-latency, real-time AI applications in various domains, such as IoT, autonomous vehicles, and smart cities. We foresee advancements in edge AI and cloud integration, which will lead to more efficient and scalable AI solutions that span across the cloud and edge devices.
    \item Green AI and Energy-efficient Computing: As the environmental impact of AI and cloud computing becomes more apparent, there will be a growing emphasis on green AI and energy-efficient computing techniques. This includes research into reducing the energy consumption of AI training and inference processes, as well as the development of more energy-efficient hardware and cloud infrastructure.
    
\end{itemize}

In conclusion, AI and cloud computing are rapidly evolving fields that offer immense potential for developing powerful and efficient AI applications. By staying abreast of the latest advancements and trends, researchers and practitioners can harness the full potential of AI in the cloud, enabling the development of cutting-edge solutions that drive innovation and improve our world.

%% file: template.bbl
\begin{thebibliography}{10}

\bibitem{7}
S.~Akter, K.~Michael, M.~R. Uddin, G.~McCarthy, and M.~Rahman.
\newblock Transforming business using digital innovations: The application of
  ai, blockchain, cloud and data analytics.
\newblock {\em Annals of Operations Research}, pp. 1--33, 2022.

\bibitem{11}
F.~Al-Turjman.
\newblock Ai-powered cloud for covid-19 and other infectious disease diagnosis.
\newblock {\em Personal and Ubiquitous Computing}, pp. 1--4, 2021.

\bibitem{2}
J.~Dodge, T.~Prewitt, R.~Tachet~des Combes, E.~Odmark, R.~Schwartz,
  E.~Strubell, A.~S. Luccioni, N.~A. Smith, N.~DeCario, and W.~Buchanan.
\newblock Measuring the carbon intensity of ai in cloud instances.
\newblock In {\em 2022 ACM Conference on Fairness, Accountability, and
  Transparency}, pp. 1877--1894, 2022.

\bibitem{15}
V.~Eramo, F.~G. Lavacca, T.~Catena, and P.~J. Perez~Salazar.
\newblock Proposal and investigation of an artificial intelligence (ai)-based
  cloud resource allocation algorithm in network function virtualization
  architectures.
\newblock {\em Future Internet}, 12(11):196, 2020.

\bibitem{4}
S.~S. Gill, S.~Tuli, M.~Xu, I.~Singh, K.~V. Singh, D.~Lindsay, S.~Tuli,
  D.~Smirnova, M.~Singh, U.~Jain, et~al.
\newblock Transformative effects of iot, blockchain and artificial intelligence
  on cloud computing: Evolution, vision, trends and open challenges.
\newblock {\em Internet of Things}, 8:100118, 2019.

\bibitem{13}
L.~Ionescu et~al.
\newblock Big data, blockchain, and artificial intelligence in cloud-based
  accounting information systems.
\newblock {\em Analysis and Metaphysics}, (18):44--49, 2019.

\bibitem{5}
S.~Juyal, S.~Sharma, and A.~S. Shukla.
\newblock Smart skin health monitoring using ai-enabled cloud-based iot.
\newblock {\em Materials Today: Proceedings}, 46:10539--10545, 2021.

\bibitem{1}
A.~Kaginalkar, S.~Kumar, P.~Gargava, and D.~Niyogi.
\newblock Review of urban computing in air quality management as smart city
  service: An integrated iot, ai, and cloud technology perspective.
\newblock {\em Urban Climate}, 39:100972, 2021.

\bibitem{9}
T.~Kurihana, E.~J. Moyer, and I.~T. Foster.
\newblock Aicca: Ai-driven cloud classification atlas.
\newblock {\em Remote Sensing}, 14(22):5690, 2022.

\bibitem{6}
M.~Li, Z.~Sun, Z.~Jiang, Z.~Tan, and J.~Chen.
\newblock A virtual reality platform for safety training in coal mines with ai
  and cloud computing.
\newblock {\em Discrete Dynamics in Nature and Society}, 2020:1--7, 2020.

\bibitem{14}
K.~N. Qureshi, G.~Jeon, and F.~Piccialli.
\newblock Anomaly detection and trust authority in artificial intelligence and
  cloud computing.
\newblock {\em Computer Networks}, 184:107647, 2021.

\bibitem{8}
A.~Salem and O.~Moselhi.
\newblock Ai-based cloud computing application for smart earthmoving
  operations.
\newblock {\em Canadian Journal of Civil Engineering}, 48(3):312--327, 2021.

\bibitem{10}
K.~K. Singh.
\newblock An artificial intelligence and cloud based collaborative platform for
  plant disease identification, tracking and forecasting for farmers.
\newblock In {\em 2018 IEEE international conference on cloud computing in
  emerging markets (CCEM)}, pp. 49--56. IEEE, 2018.

\bibitem{3}
J.~Wan, J.~Yang, Z.~Wang, and Q.~Hua.
\newblock Artificial intelligence for cloud-assisted smart factory.
\newblock {\em IEEE Access}, 6:55419--55430, 2018.

\bibitem{12}
Z.~Wang, Z.~Zhou, H.~Zhang, G.~Zhang, H.~Ding, and A.~Farouk.
\newblock Ai-based cloud-edge-device collaboration in 6g space-air-ground
  integrated power iot.
\newblock {\em IEEE Wireless Communications}, 29(1):16--23, 2022.

\end{thebibliography}
